\documentclass[letterpaper, 10 pt, conference]{ieeeconf}  

\IEEEoverridecommandlockouts                              

\usepackage[hyphens]{url}
\usepackage{hyperref}
\usepackage[ruled,norelsize,linesnumbered]{algorithm2e}
\usepackage{amsmath}
\usepackage{listings}
\usepackage{amssymb}
\usepackage{graphicx}
\usepackage{color}
\usepackage{caption}
\usepackage{multirow}
\usepackage{subcaption}
\newsavebox{\measurebox}

\usepackage{mathtools}

\newcommand\eat[1]{}

\usepackage{color,soul}

\IEEEoverridecommandlockouts                              

\renewcommand{\baselinestretch}{0.97}

\makeatletter
\def\ps@titlepagestyle{%
  \def\@oddfoot{\mycopyrightnotice}%
  \def\@evenfoot{}%
}
\def\mycopyrightnotice{%
  \begin{minipage}[\textheight]{\textwidth}  
   \footnotesize \textbf  { \hrule \smallskip
  \copyright~2021 IEEE. Personal use of this material is permitted.  Permission from IEEE must be obtained for all other uses, in any current or future media, including reprinting/republishing this material for advertising or promotional purposes, creating new collective works, for resale or redistribution to servers or lists, or reuse of any copyrighted component of this work in other works.}
  \end{minipage}
}
\makeatother

\begin{document}
\bstctlcite{IEEEexample:BSTcontrol}

\title{\LARGE \bf
SQRP: Sensing Quality-aware Robot Programming System for Non-expert Programmers
}



\author{Yi-Hsuan Hsieh$^{1}$, Pei-Chi Huang$^{2}$, and Aloysius K Mok$^{1}$
\thanks{*This work is partially supported by the Office of Naval Research under ONR Award N00014-17-2216}
\thanks{$^{1}$Yi-Hsuan Hsieh and Aloysius K Mok are with Department of Computer Science, University of Texas at Austin,
        {\tt\small yihsuan, mok@cs.utexas.edu}}%
\thanks{$^{2}$Department of Computer Science, University of Nebraska Omaha, 
    {\tt\small phuang@unomaha.edu}}%
}

\maketitle

\begin{abstract} 
Robot programming typically makes use of a set of mechanical skills that is acquired by machine learning. Because there is in general no guarantee that machine learning produces robot programs that are free of surprising behavior, the safe execution of a robot program must utilize monitoring modules that take sensor data as inputs in real time to ensure the correctness of the skill execution. Owing to the fact that sensors and monitoring algorithms are usually subject to physical restrictions and that effective robot programming is sensitive to the selection of skill parameters, these considerations may lead to different sensor input qualities such as the view coverage of a vision system that determines whether a skill can be successfully deployed in performing a task. Choosing improper skill parameters may cause the monitoring modules to delay or miss the detection of important events such as a mechanical failure. These failures may reduce the throughput in robotic manufacturing and could even cause a destructive system crash. To address above issues, we propose a sensing quality-aware robot programming system that automatically computes the sensing qualities as a function of the robot's environment and uses the information to guide non-expert users to select proper skill parameters in the programming phase. We demonstrate our system framework on a 6DOF robot arm for an object pick-up task. 
\end{abstract}

\section{Introduction}\label{sec:intro}
Skill-based robot programming that composes a set of low-level skills into a high-level capability has been widely used in robotic manufacturing systems because of the need for reusability~\cite{huang2018skill}. Successful execution of a skill requires real-time sensor inputs for monitoring the correctness of the skill execution. One popular sensing method is to use cameras to provide different views to cover certain critical aspects in a skill execution. There are several technical challenges to this approach. First, sensors have their own coverage limitations, such as a camera's limited field of view, object occlusions in the work environment and also the physical requirements imposed by the detection algorithm. Second, a robot skill usually requires the proper setting of the skill parameters to achieve the task goal. Without sufficient camera coverage, a robot system may miss a crucial deadline in the detection of an execution failure that results in reduced system performance or even a catastrophic system crash. Some extant work assumes that there are enough sensors to achieve the monitoring requirements~\cite{inceoglu2018failure} while other works focus on reconfiguring the cameras to meet the goals of the robotic tasks~\cite{hanoun2016target}. However, it requires time and expertise for performing reconfiguration, and we do not want to reconfigure the cameras if it can be avoided. These are important issues if robot programming is to be made accessible to non-expert programmers who need to know if the current camera settings can or cannot support a robotic skill. This paper is a first step towards treating these issues by providing meaningful feedback to the programmer that quantifies the task-effectiveness of the chosen system parameters such as the adequacy in camera coverage. To address the above issues, we propose a sensing quality-aware robot programming system we name SQRP that incorporates explicit sensing requirements in the skill definition. We include temporal sensing requirements in Metric Temporal Logic (MTL)~\cite{koymans1990specifying} formulas that prescribe what to monitor and when to monitor. We also include spatial sensing requirements that prescribe where to monitor. In the programming phase, our system examines the sensing requirements to determine if the current system configurations and camera settings can support a robotic skill and to guide the programmer to choose the proper skill parameters based on the sensing quality. 

\begin{figure}[t!]
\centering
\includegraphics[width=0.48\textwidth]{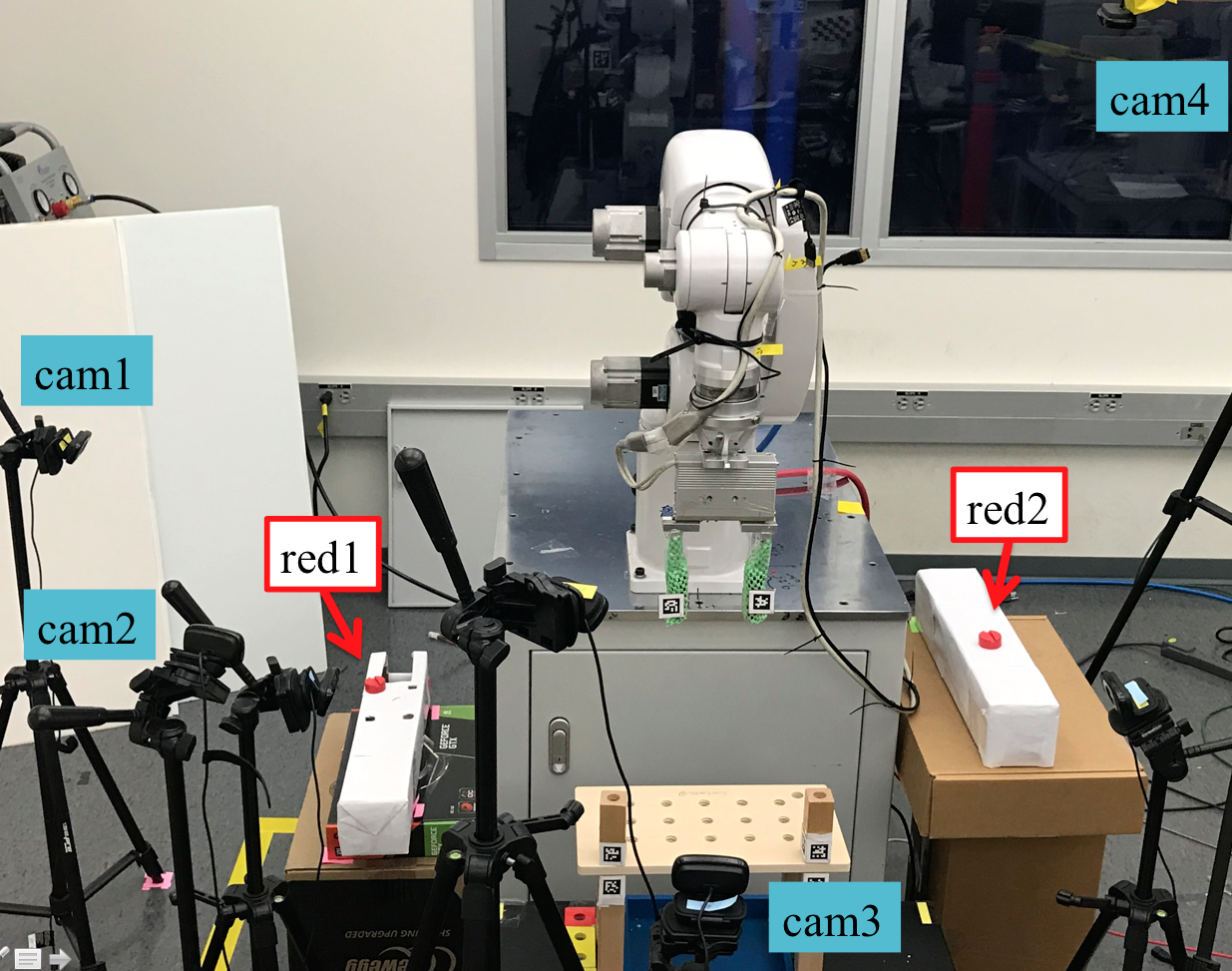}
\caption{Real world environment.}
\label{fig:realworld_scene}
\vspace{-15pt}
\end{figure}

The contributions of this paper are twofold. Firstly, we introduce the sensing requirements in the robot skill that includes both temporal and spatial sensing requirements. Secondly, based on these requirements, we compute the sensing qualities in the programming phase to assist users to choose the proper skill parameters. In our experimental evaluation, we show the benefit of exposing sensing quality in the programming phase as it assists users to choose a proper set of skill parameters to reduce the execution time of a robotic task, especially when a fault occurs during execution which may require a sub-task to be redone. We use a 6DOF robot arm to demonstrate the application of one of its skill sets - ``Pickup'' skill, both in the simulation and the real world environment, as shown in Figure~\ref{fig:realworld_scene}.

\section{System Overview}\label{sec:overview}
\begin{figure}
\centering
\includegraphics[width=0.49\textwidth]{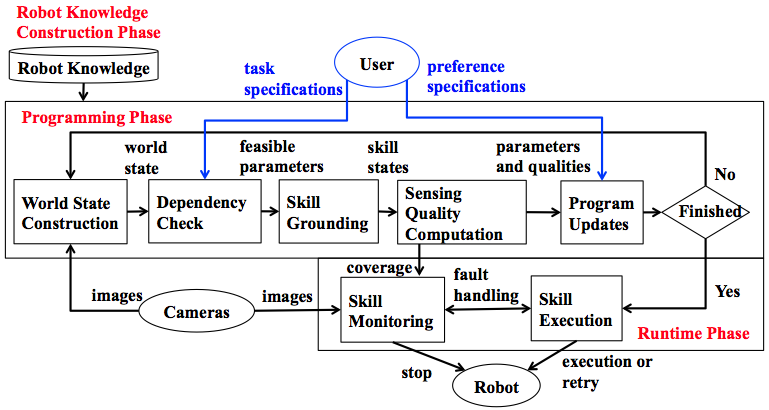}
\caption{The overview of the SQRP.}
\label{fig:sys_overview}
\vspace{-15pt}
\end{figure}

Building on our previous work~\cite{hsieh2019lasso}, we develop our robot programming system in three phases, as shown in Figure~\ref{fig:sys_overview}. First, in the \textit{Robot Knowledge Construction Phase}, a robot knowledge base is constructed by system developers which will be automatically adapted for the the actual operating environment. Then, in the \textit{Programming Phase}, the user programmer specifies the tasks to be performed with the selected robot skills and a set of parameters. Our system checks if the specifications are supported by the robot system and then computes the corresponding sensing qualities for the chosen parameter set which is the feedback to guide the application programmer. In the subsequent \textit{Runtime Phase}, the robot executes the task by deploying the selected skills while our system monitors the correctness of the robot's execution. If faults are detected, the system fault handler will determine the response. 

\section{Skill Definition and Requirement Specification}\label{sec:skilldef}
The section describes how to defines a robot skill with the corresponding sensing requirements.

\vspace{-3pt} 
\subsection{Skill Definition}\label{subsec:skill_def}
A skill is specified in the planning domain definition language (PDDL)~\cite{mcdermott1998pddl} which describes each skill's preconditions and postconditions. A skill $S_i$ consists of $n_i$ number of sequential primitive actions as follows: $S_i = <a_1, a_2, \dots, a_n>$. Each primitive action is further annotated with a symbolic execution time interval $[t_{s}, t_{e}]$, where $t_{s}$ and $t_{e}$ are the start time and end time respectively. These two parameters will be grounded in the programming phase based on the target robot system and the target work environment. 

\subsection{Sensing Requirements}\label{subsec:sense_require}
The sensing requirements of a skill define what/when/where the targets have to be covered by the robotic system sensors in order to determine the applicability of the skill. There are two types of sensing requirements: temporal sensing requirements and spatial sensing requirements.

\subsubsection{Temporal Sensing Requirements}
Temporal sensing requirements define the desired targets, when and how long that the targets have to be covered by the sensors. In this paper, we define our skill by using Metric Temporal Logic (MTL)~\cite{koymans1990specifying}, a logic specification formalism that is used to specify a temporal property in the context of time intervals. The syntax of a MTL formula $\varphi$ is defined as follows:
\begin{gather*}
\varphi ::= a \; | \; \neg \varphi  \; | \; \varphi \wedge \varphi \; | \; \square_I \varphi \; | \; \Diamond_I \varphi
\end{gather*}
  \normalsize
where $a \in A$ and $A$ is a set of atomic propositions. The set of atomic propositions of a skill is obtained from the grounded predicates of the skill's preconditions and postconditions in the programming phase. The temporal operator $\square$ and $\Diamond$ are the ``always'' and ``eventually'' modal operators respectively. The $I \in \mathbb{R}_{\geq0}\times\mathbb{R}_{\geq0}$ defines the time interval of the temporal operator. In our system, the time interval $I$ is delimited by the start time and the end time of the primitive actions. Formula $\square_I \varphi$ is true {\em iff} $\varphi$ is always true in the time interval $I$. The $\square$ operator defines the persistent states of grounded predicates; we use it to specify the correctness criterion of the skill execution. For instance, the following MTL formula $\square_{[a_1.t_s, a_2.t_e]} (open)$ specifies that the robot gripper remains open during the time interval $[a_1.t_s, a_2.t_e]$. On the other hand, formula $\Diamond_I \varphi$ is true {\em iff} $\varphi$ is true sometime in the time interval $I$. This specification is useful to describe a runtime event, such as a runtime fault that the system needs to focus on. For example, $\Diamond_{[a_4.t_s, a_4.t_e]} \; (obj\_on\_table \wedge \; \neg \; open \wedge \; \neg hold)$ specifies the state after the runtime fault event ``object mistakenly slip'' occurs. It specifies that at some time between $a_4.t_s$ and $a_4.t_e$ a slip fault occurs; the object is (still) on the table, the robot gripper is not open and the gripper does not hold anything. 

To determine the satisfaction of the MTL formulas at runtime, we need to provide adequate sensing capability in hardware and software. The grounded predicates that are specified in the MTL formulas concern the target objects that have to be within the sensing coverage of the sensors. The time interval specified in the MTL formulas prescribes when and for how long the targets have to be monitored by the sensors. 

\subsubsection{Spatial Sensing Requirements}
Spatial sensing requirements prescribe where each of the target, the grounded predicate mentioned in the MTL formulas, has to be monitored by the sensors. We denote a target by $P_i$, where $1 \leq i \leq N$ and the $N$ is the number of the total targets. For each target, we define a set of 3D bounding boxes $Box^{f}_{i}$ that together enclose the physical objects that make up the target $P_i$. One can reduce one of the dimensions of a 3D bounding box to form a 2D bounding box based on the application's need. As an example, for the literal $open$, our system defines a set of two bounding boxes that enclose two ArUco markers at the tip of the robot gripper, as shown in Figure~\ref{fig:occlusion_realworld} (a). Each bounding box $box^{f}_{i,j} \in Box^{f}_{i}$ is represented as eight vertices in the Cartesian coordinates relative to the $f$ coordinate frame, where $1 \leq j \leq b$ and the $b$ is the number of bounding boxes in the set $Box^{f}_{i}$. We note that some coordinate frames may change their 3D locations relative to the world coordinate as time progresses.

We require that $box^{f}_{i,j}$ meets the following two spatial sensing requirements: (1) It is covered inside the sensing range of the sensors, in our case the camera viewing frustum~\cite{hanoun2016target}; (2) It is covered inside the detection range of the object detection algorithm. For accurate object detection, we need the target to be within a range of some nominal distances from the camera and the pose of the target to be within a certain tolerance of some nominal poses.

\section{Sensing Quality-aware Robot Programming (SQRP) System}\label{sec:method}

\subsection{Robot Knowledge Construction Phase}\label{subsec:robot_knowledge_construction}
To equip our robot system with the knowledge of its surroundings, the system developers define four types of information as follows: 1) Sensing knowledge: each camera that monitors the work space is calibrated; 2) Object appearances and locations: each object that can be manipulated in the work space is represented by its name, type (e.g., colors), size (e.g., volume of a 3D bounding box) and its 3D location if available; 3) Robotic arm capability: the arm's spatial reachability subspace, motion profile and its forward/inverse kinematics; and 4) Skill requirements: the skills mentioned in Section~\ref{sec:skilldef} that can be used by the application programmer.

\subsection{Programming Phase} \label{subsec:program_phrase}
\subsubsection{World State Construction}
The world state at time $t$ is expressed as a conjunction of $n$ literals, $W = \wedge^{i=n}_{i=1} L_{i}$. For the initial state, the truth values of all literals are determined by the sensor inputs. Here, we assume that all literals that are not defined by sensor input in the initial state are assumed to be false. 

\subsubsection{Dependency Check}
First, our system takes the user task specifications as inputs to check if there exist other skill parameters that satisfy the user inputs. For instance, given {\em Pickup red screw} as the user's input, our system includes all the red screws that are in the work environment as the possible values of the skill parameters. Our system then performs a dependency check to see if the world state meets the preconditions of the chosen skill. If not, we will apply the Fast-forward solver~\cite{hoffmann2001ff} to generate execution plans as suggestions to the user. Our systems also checks whether our robot can reach the target or not and only outputs the feasible parameters. If the specified skill and the corresponding skill parameters do not pass the check, our system will not allow the user to proceed. 

\subsubsection{Skill Grounding}
For each skill parameter, we ground the skill definition and the information from the robot's knowledge base to the target robot system. We first obtain the overall execution time of the skill with the specified skill parameters by checking the S-curve motion profile for our real-world robot and by using the time parameterization algorithm from Moveit!~\cite{chitta2012moveit} for our simulation robot. From the MTL formulas, we obtain $T^{p}$, the total execution time that spans the robot's motion trajectory. We then sample the robot's state as it moves along the trajectory. With respect to a skill parameter $p$ and sample time $t_j$, we define a skill state, denoted by $S^{p}_{t_j}$ to be the conjunctive form $\wedge^{i=n_j}_{i=1} L^{p}_{i,t_j}$, where $n_j$ is the number of conjunction literals of each time sample $t_j$. We obtain $n_j$ from the MTL formulas in the skill definition. Depending on the sampling method, the temporal distance between two samples in the trajectory may not be constant. For a given trajectory, we have a sequence of skill states $S^{p}_{t_1}, S^{p}_{t_2}, \dots S^{p}_{t_m}$, where $m$ is the number of samples of the trajectory.  

\subsubsection{Sensing Quality Computation}
We provide two metrics to define sensing quality. The first metric is called ``overall average sensing coverage'', which defines the percentage of camera coverage of the entire skill. A higher value implies that more time intervals of the skill execution are monitored by cameras. Accordingly, there is a higher chance to capture events of concerns in time, such as runtime faults that may not be explicitly specified by application developer. However, some runtime faults tend to happen in a specific time interval. For instance, for the ``Pickup'' skill that is used to lift a target object up and move it to another location, the target usually slips during the ``lift up'' process instead of while moving to another location. Thus a high overall coverage does not necessary guarantee that the critical time period of the ``lift up'' process will be adequately monitored. Without monitoring this time interval, the pertinent runtime fault may not be detected fast enough. 

To address the above issue, we introduce the second metric, the ``event of interest average sensing coverage'', which computes the average coverage for all the time intervals that are spanned by the interval arguments of all the $\Diamond$ modal operators that appear in the MTL formulas in the skill definition. 

To compute the two metrics, we first define the concept of sensing coverage. We say that camera $c_x$ covers a literal $L^{p}_{i,t_j}$ if the camera meets the sensing requirements that are sufficient to determine the truth value of the literal. We define the coverage of a literal $L^{p}_{i,t_j}$ in a skill state by camera $c_x$ to be

\vspace{-5pt}
\small
\begin{gather*}
    C(L^{p}_{i,t_j}, c_{x})=
    \begin{cases}
      1, & \text{if $c_{x}$ covers $L^{p}_{i,t_j}$} \\
      0, & \text{otherwise}
    \end{cases}
  \end{gather*}
      \normalsize
For a given set of cameras, the coverage of the literal $L^{p}_{i,t_j}$ is defined as 

\vspace{-5pt}
  \small
\begin{gather*}
    C(L^{p}_{i,t_j})=
    \begin{cases}
      1, & \text{if $\sum_{x=1}^{\omega} C(L^{p}_{i,t_j}, c_{x}) \geq k$} \\
      0, & \text{otherwise}
    \end{cases}
  \end{gather*}
    \normalsize
 where $k$ is the minimum number of cameras that are required to determine the truth of the literal, and $\omega$ is the number of the camera in the set. For a skill state, denoted by $S^{p}_{t_j}$, we define the coverage of the skill state as
 
 \vspace{-5pt}
  \small
  \begin{gather*}
    C(S^{p}_{t_j})=
    \begin{cases}
      True, & \text{if $C(L^{p}_{i,t_j}) = 1$, $\forall L^{p}_{i,t_j}$ in $S^{p}_{t_j}$ } \\
      False, & \text{otherwise}
    \end{cases}
  \end{gather*}
  \normalsize
We say that a skill state is covered if all its literals are covered a set of cameras. 

We now define the first metric of the skill parameter $p$ to be $Q^{p}_{avg}$. Suppose $SecT^{p}_\gamma$ is a sequence of time points $<t_j, t_{j+1}, \dots, t_{j+n_{\gamma}}>$ such that all the skill states represented by these time points are all covered. In other words, the conjunction $C(S^{p}_{t_j}) \wedge \dots \wedge C(S^{p}_{t_{j+n_{\gamma}}})$ is $true$, where $n_{\gamma}+1$ is the number of the trajectory sample points and $1 \leq \gamma \leq \Gamma$, where $\Gamma$ is the number of such time segments in the trajectory. Note that we have either $C(S^{p}_{t_{j-1}})$ is $false$ or $t_j = t_1$, the start of the trajectory. Also, we have either $C(S^{p}_{t_{j+n_{\gamma}+1}})$ is $false$ or $t_{j+n_{\gamma}} = t_m$, the end of the trajectory. We  define $\Delta SecT^{p}_\gamma = (t_{j+n_{\gamma}} - t_{j})$ to be the length of the time interval of $SecT^{p}_\gamma$. Finally, $Q^{p}_{avg}$ is defined as

 \vspace{-5pt}
\small
  \begin{gather*}
   Q^{p}_{avg} =  {\frac{\sum_{\gamma=1}^{\gamma=\Gamma} \Delta SecT^{p}_\gamma }{T^{p}}}
  \end{gather*}
    \normalsize

We define the second metric of the skill parameter $p$ to be $Q^{p}_{eoi}$. The length of the time intervals that are spanned by all the $\Diamond$ modal operators is denoted by $T^{p}_{eoi}$. Similar to computing $Q^{p}_{avg}$, we use $SecT^{p}_{eoi,\gamma}$ to denote a sequence of time points in $T^{p}_{eoi}$ where the skill states are covered. We use $\Delta SecT^{p}_{eoi,\gamma}$ to denote the time duration of $SecT^{p}_{eoi,\gamma}$. With above notations we define  

 \vspace{-5pt}
\small
  \begin{gather*}
   Q^{p}_{eoi} =  {\frac{\sum_{\gamma=1}^{\gamma=\Gamma} \Delta SecT^{p}_{eoi,\gamma} }{T^{p}_{eoi}}}
  \end{gather*}
    \normalsize
where $\Gamma$ is the total number of such time segments within $T^{p}_{eoi}$. 

\subsubsection{Preference Specifications and Program Updates}
A programmer may determine what skill parameters s/he wants or based on the following criteria: $p* = arg\max\limits_{p} Q^{p}_{avg}$ or $p* = arg\max\limits_{p} Q^{p}_{eoi}$. After determining the skill parameters, our programming system saves the specified skill and updates the world state based on the skill's postcondition to allow the programmer to specify the next skill to be deployed.

\vspace{-5pt} 
\subsection{Runtime Phase}\label{subsec:runtim_phase}
\vspace{-2pt} 
After all the skill specifications are completed, our system is ready to execute the skills. Concurrently, the skill monitoring module which consists of several parallel processes takes the camera images and coverage information as input to determine the result of the skill execution. When an execution fault is detected, a fault handler in our SQRP system will determine the proper responses, such as stopping the robot and performing a retry.
\section{Performance Evaluation}\label{sec:experiment} 
This section describes the experimental evaluation and discusses the performance efficiency of our SQRP system. We use the ``Pickup'' skill with robot suction in the simulation environment and we use the robot two-finger gripper in our real-world experiment to pick up an object as the scenario of our experiment.

\subsection{Simulation-based Experiments}\label{subsec:simulatin_exp}

\begin{figure}[t!]
\centering
\includegraphics[height = 35mm]{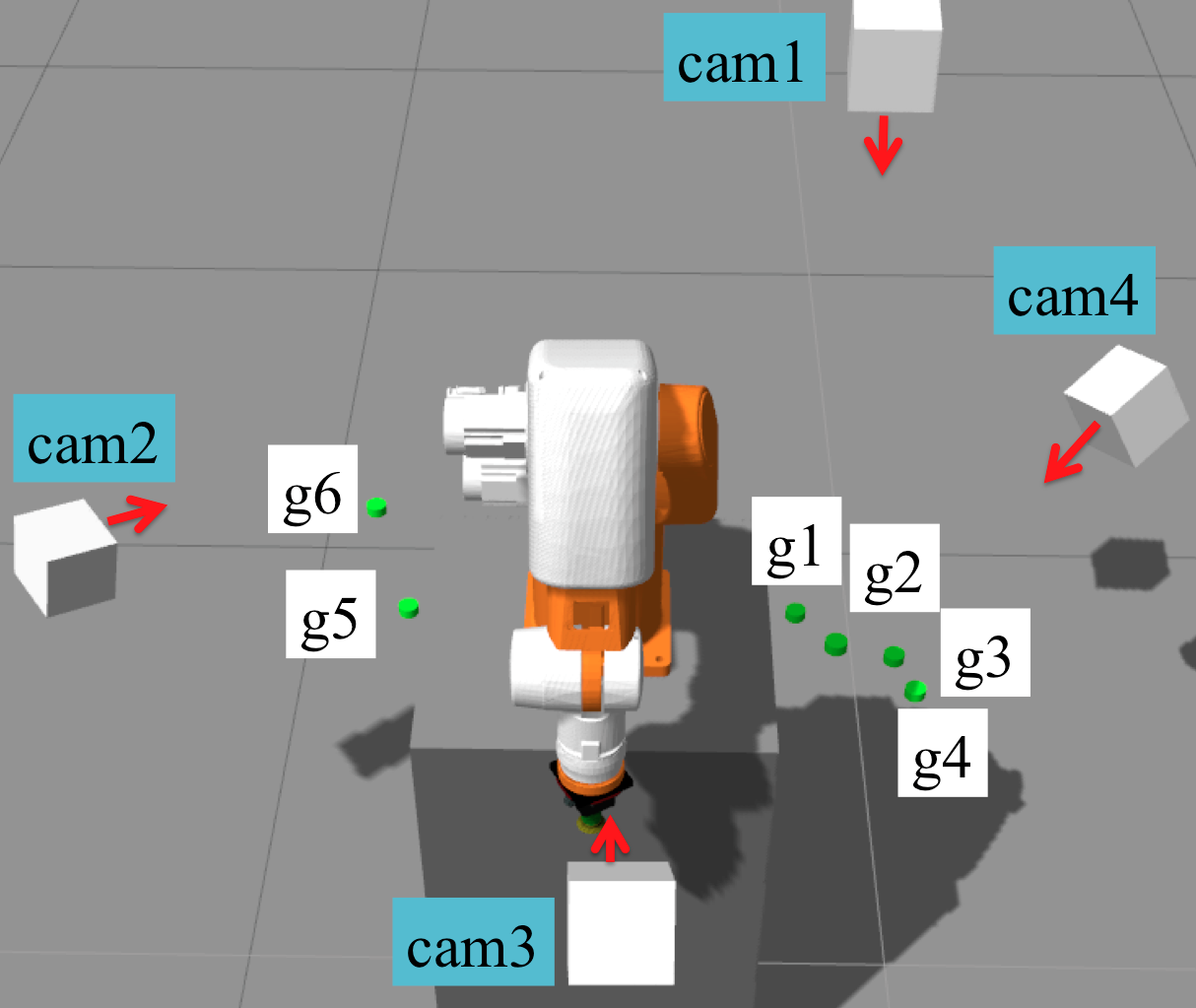}
\hspace{5pt}
\includegraphics[height = 35mm]{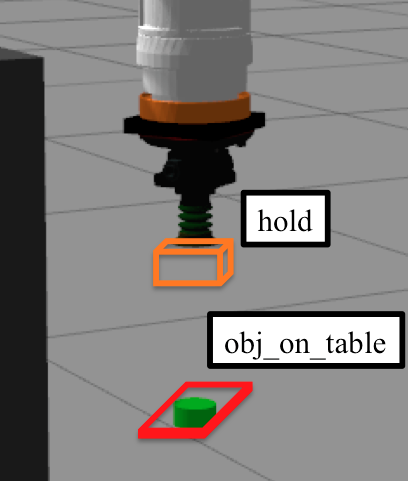} 
\caption{Simulation scene: (a) Environment (b) Spatial sensing requirements as illustrated in orange and red boxes.}
\label{fig:simulation_two}
\vspace{-10pt}
\end{figure}

\begin{figure}[ht!]
\centering
\includegraphics[width=0.15\textwidth]{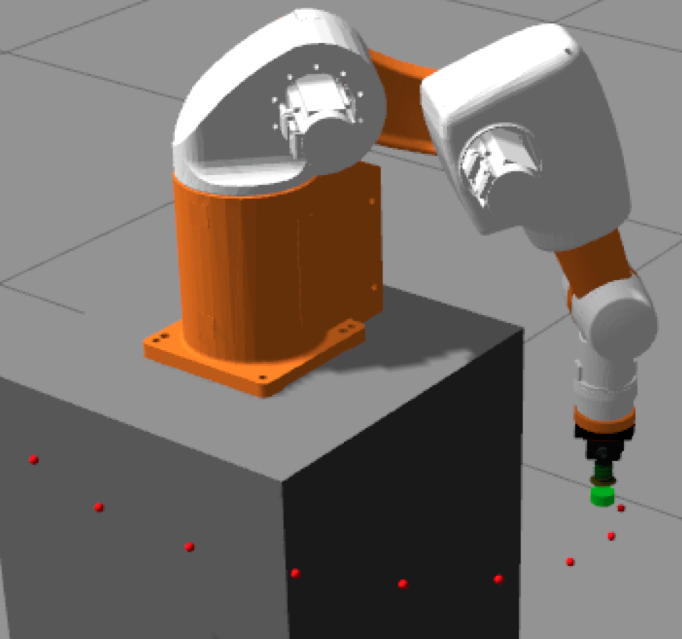}
\includegraphics[width=0.15\textwidth]{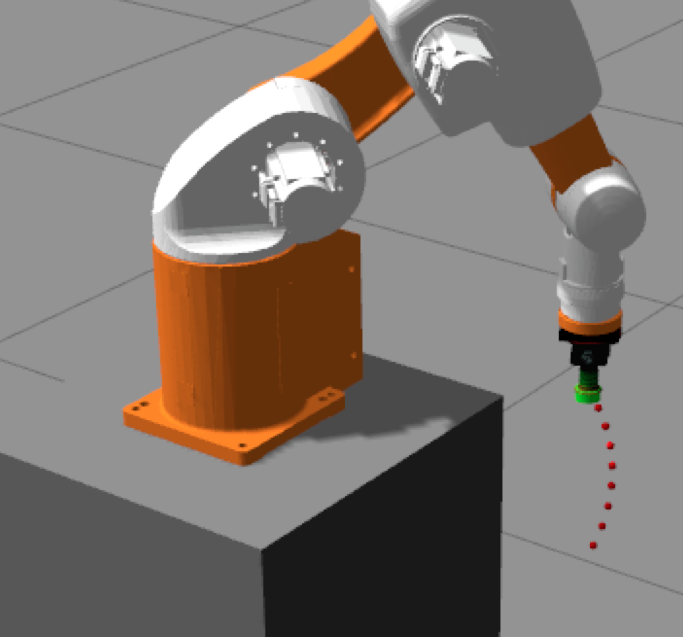}
\includegraphics[width=0.15\textwidth]{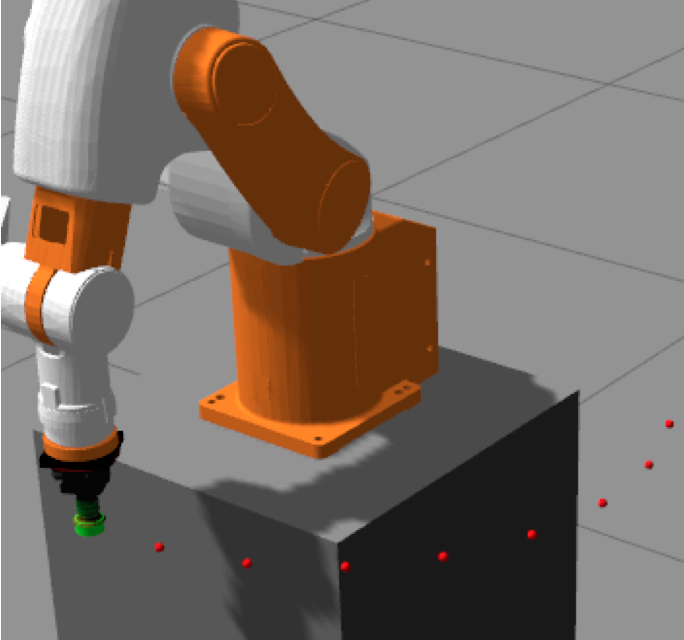}
\caption{Primitive actions. (a) Move above the target. (b) Lift the object up. (c) Move back to initial pose.}
\label{fig:simu_traj}
\vspace{-12pt}
\end{figure}

\subsubsection{Experimental Setup and Deployment}
The simulation environment, as shown in Figure~\ref{fig:simulation_two} (a) is implemented in the Gazebo simulator~\cite{koenig2004design} with a 6DOF robot arm equipped with a suction cup at its end-effecter. Our system is built on top of the Robot Operating System (ROS)~\cite{quigley2009ros} that communicates with the simulation camera sensors and the robot simulation system. Moveit!~\cite{chitta2012moveit} is used for robot planning and collision detection.

Six cameras are placed in locations to highlight the differences and benefits of using different sensing quality metrics. There are six green objects that can be selected as the skill parameter for the ``Pickup'' skill. $g1$ and $g5$ are deployed at the locations that take similar execution time, while the remaining green objects are randomly generated with the constraint that they are able to be seen by one of the cameras.

The ``Pickup'' skill consists of five primitive actions $<a1, \dots, a5>$ with the following motion preferences. $a1$: From the initial pose, Figure~\ref{fig:simulation_two} (a), the robot arm moves to two centimeters above the target, Figure~\ref{fig:simu_traj} (a). $a2$: The arm moves down for two centimeters. $a3$: suction. $a4$: The arm lifts itself up, Figure~\ref{fig:simu_traj} (b). $a5$: it moves back to the initial position, Figure~\ref{fig:simu_traj} (c). 

The temporal sensing requirements are obtained from the following MTL formulas: $\square_{[a_1.t_s, a_2.t_e]} \; (\neg \; hold \; \wedge \; obj\_on\_table) \wedge \square_{[a_4.t_s, a_5.t_e]} \; (hold \; \wedge \; \neg \; obj\_on\_table)$. A runtime fault, object-mistakenly-slips, is defined as follows: $\Diamond_{[a_4.t_s, a_4.t_e]} \; (\neg hold \wedge obj\_on\_table)$. The $hold$ and $obj\_on\_table$ prescribe that the object is stuck to the suction cup and the target green object is on the table respectively. The spatial sensing requirements are shown in Figure~\ref{fig:simulation_two} (b). The skill states are sampled based on the waypoints generated from the Moveit!~\cite{chitta2012moveit}'s planner. Our system then computes two sensing quality metrics for each of the green object.

\subsubsection{Performance of Computing Sensing Quality}

\begin{figure}[t!]
\centering
\includegraphics[width=0.49\textwidth]{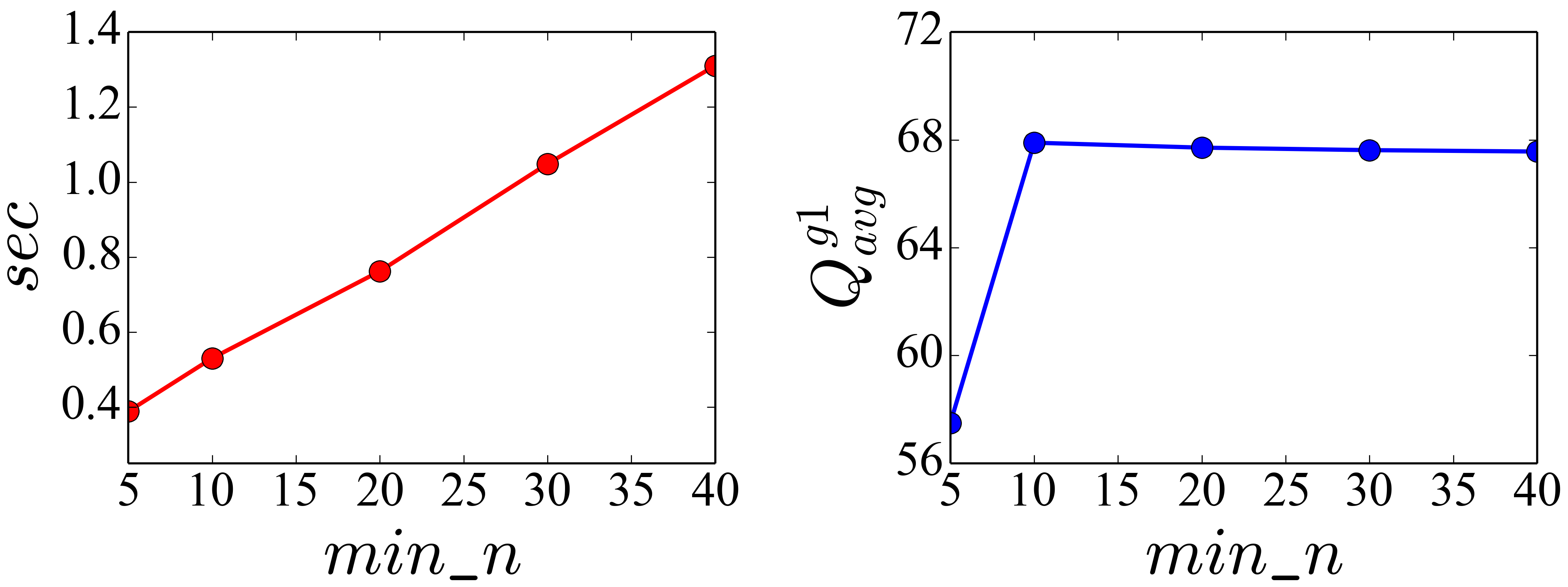}
\caption{Results for using different $min\_n$. (a) Execution time v.s. $min\_n$. (b) $Q^{g1}_{avg}$ v.s. $min\_n$.}
\label{fig:exp1_acct}
\vspace{-10pt}
\end{figure}

\begin{figure}[t!]
\centering
\includegraphics[width=0.49\textwidth]{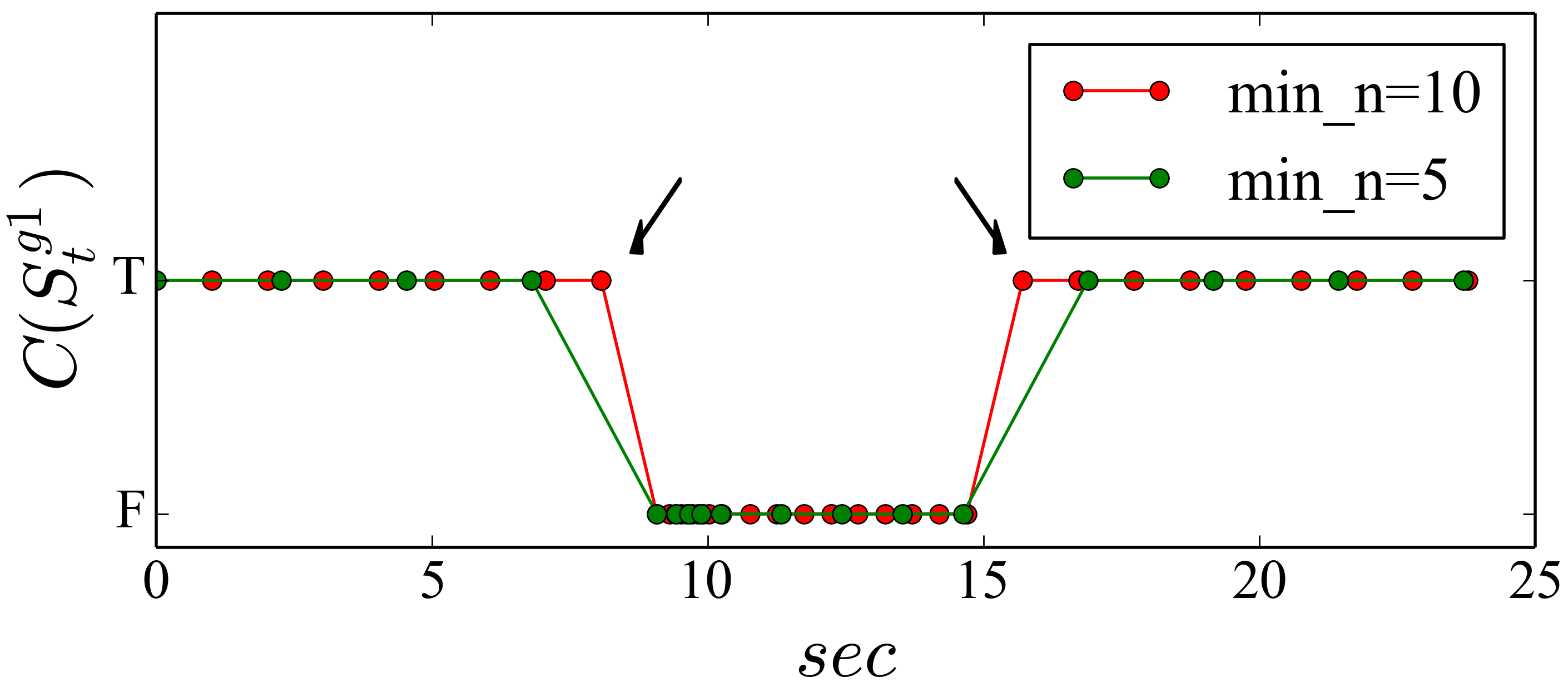}
\caption{Different $C(S^{g1}_{t})$ when using different $min\_n$. T and F represent \textit{Truth} and \textit{False} respectively.}
\label{fig:exp1_n5n10}
\vspace{-10pt}
\end{figure}

Since our sensing quality metrics are computed based on skill states that are sampled in discrete time, we first evaluate the performance and the ``overall average sensing quality'' when using different numbers of skill states to pick up $g1$. To get a different number of skill states, we set the minimum number of waypoints, $min\_n$, in Moveit!~\cite{chitta2012moveit}'s planner from $5$ to $40$ for each primitive action except, $a3$. In Figure~\ref{fig:exp1_acct} (a), $x$ axis shows $min\_n$ and $y$ axis shows the execution time to compute $Q^{g1}_{avg}$. The computation time increases from 0.388 sec to 1.31 sec as the $min\_n$ increase from 5 to 40, as the number of the states that need to be evaluated increase. In Figure~\ref{fig:exp1_acct} (b), $x$ axis shows $min\_n$ and $y$ axis shows the value of $Q^{g1}_{avg}$. We can see a convergence to within 67.6\% to 67.9\% for $min\_n$ values 10, 20, 30, and 40. The intuition here is that with a bigger number of skill states, the discrete time simulation approaches continuous time motion and therefore outputs similar results. However, with insufficient number of skill states, $min\_n = 5$, we only get 57.5\%, since our system misses significant coverage information, as shown in Figure~\ref{fig:exp1_n5n10}, where the $x$ axis represents time in seconds, and $y$ axis represents the skill coverage, $C(S^{g1}_{t})$. The areas that are pointed to by two black arrows show that when using $min\_n=5$, there is no skill state that can be evaluated at around 8.1 and 15.7 seconds, whereas there are such states for $min\_n=10$. Since we compute $\Delta SecT^{g1}_{r}$ more conservatively and only compute it when the covered skill states are sequentially covered, the lack of skill states results in less coverage in the computation. Different values of $min\_n$ that are required depend on the robot's motion and the relative location between the targets and the cameras. Since our experiment scenarios involve motion so as to picking $g1$, we choose 10 points as our $min\_n$ for the rest of experiments.

\subsubsection{Sensing Quality Results}
\begin{figure}[t!]
\centering
\includegraphics[height = 26mm]{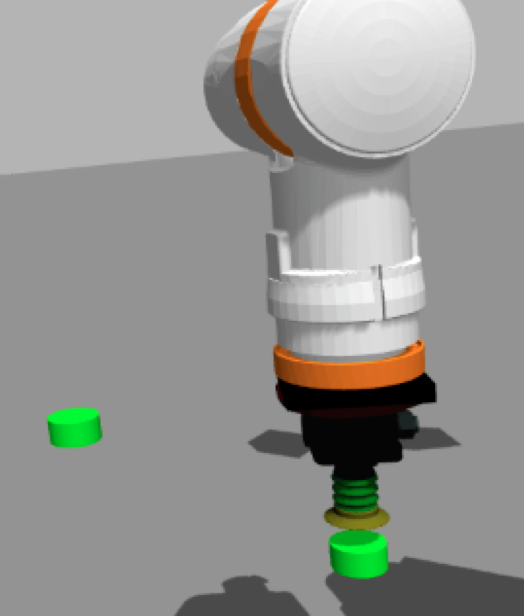} 
\hspace{10pt}
\includegraphics[height = 26mm]{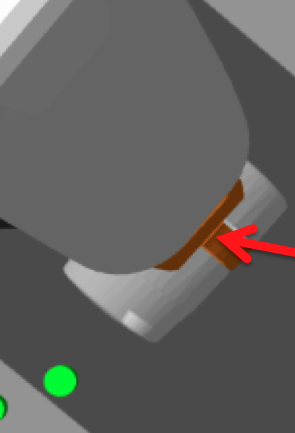}
\caption{Simulation experimental scene: (a) $g5$ in $cam2$'s perspective. (b) $g1$ is occluded from the perspective of $cam1$.}
\label{fig:robot_occ}
\vspace{-12pt}
\end{figure}

The results of sensing quality and the skill execution time for each of the green object are shown in Table~\ref{tab:sensing_qs}. Objects $g1$ and $g5$ take almost the same time for execution. However, there is a tradeoff between two sensing quality metrics. Even though choosing $g5$ results in lower $Q^{g5}_{avg}$, it still has 100\% of $Q^{g5}_{eoi}$ owing to the coverage from $cam2$, as shown in Figure~\ref{fig:robot_occ} (a). If the pre-specified runtime fault occurs, e.g., the object mistakenly slips, our system can provide fast detection time. However, if a fault occurs outside the pre-specified runtime fault interval, our system may not be able to capture it in time. On the other hand, for selecting $g1$, even though $Q^{g1}_{avg}$ is up to 67.80\%, $Q^{g1}_{eoi}$ is 0\% owing to the occlusion from the robot arm, as shown in Figure~\ref{fig:robot_occ} (b). If the pre-specified runtime fault occurs, our system may delay to find out. 

For object $g6$, even though it can be seen by the camera $cam2$, it is too far for the robot arm to reach. Therefore, it fails at the dependency check module. For objects $g3$ and $g4$, both have good overall sensing coverage but the ``event of interest average sensing quality'' are lower than $g2$. According to Table~\ref{tab:sensing_qs}, $g2$ has the shortest execution time, large ``overall average sensing quality'' 80.25\% and 100\% of the ``event of interest average sensing quality''. If the user selects $g2$ as the target object, our system will obtain better execution time and sensing coverage. Therefore, the best option may be to pick up $g2$.

\begin{table}[h!]
\caption{Sensing Quality for Green Objects}
\label{tab:sensing_qs}
\centering
\begin{tabular}{ | l | l | l | l | l | l | l |}
    \hline
    Metric & $g1$ & $g2$ & $g3$ & $g4$ & $g5$ & $g6$ \\ \hline
    $Q^{g_i}_{avg} (\%)$  & 67.80 &  80.25 & 83.87 & 79.88 & 36.67 & N/A\\ \hline
    $Q^{g_i}_{eoi} (\%)$ & 0.0 &  100.0 & 66.12 &  55.41 & 100.0 & N/A\\ \hline
    $Time (sec)$  & 23.81 &  13.83 & 19.62 &  18.78 & 23.89 & N/A\\ \hline
    \end{tabular}
\end{table}

\vspace{-6pt}
\subsection{Real-World Experiments}\label{subsec:real-world}

\subsubsection{Experimental Setup and deployment}
In the real-world experiments, we demonstrate the benefit of selecting a proper skill parameter when a pre-specified runtime fault, object-mistakenly-slips occurs and requires the system to perform a retry that results in a shorter skill execution time. We select the ``Pickup'' skill for the case where with the object to be lifted up may slip out of the gripper as a fault. The environment setup is shown in Figure~\ref{fig:realworld_scene}. Four cameras, $cam1, \dots, cam4$, are used and are calibrated with the robot's coordinates. There are two identical red screws, $red1$ and $red2$ that can be selected to be picked up. The robot motion is similar to the simulation setup. The temporal sensing requirements are similar to the simulation with an additional specification: $\square_{[a_1.t_s, a_2.t_e]} \; (open) \wedge \square_{[a_4.t_s, a_5.t_e]} \; (\neg open)$. The spatial sensing requirements are shown in Figure~\ref{fig:occlusion_realworld} (a). The spatial sensing requirements of the literal $open$ and the literal $hold$ both are defined by two markers, illustrated in orange boxes, but the literal $hold$ also includes the blue bounding box. The sensing requirements of the literal $obj\_on\_table$ defines as a box that encloses the red screw, shown in the red box. To generate the skill states, we use a S-curve motion profile and sample each primitive action with equal-length time intervals $1$ second long in our experiment except the end of each primitive action. In this experiment setup, we only consider the occlusion from the robot arm. To mainly focus on sensing quality measurements, we use relatively simple detection methods, such as ArUco marker detection~\cite{opencvaruco} and color detection. 

\subsubsection{Experimental Results}
\begin{figure}[t!]
\centering
\includegraphics[height=28mm]{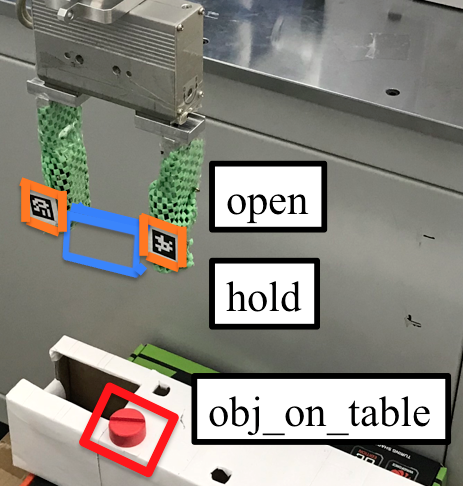}
\includegraphics[height=28mm]{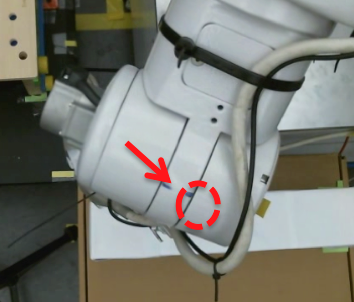}
\includegraphics[height=28mm]{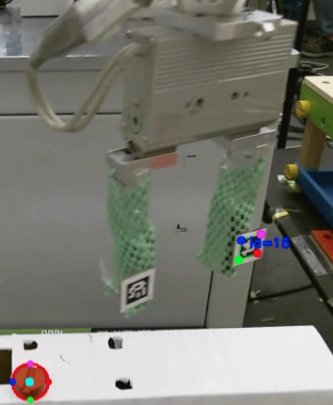}
\caption{Real-world experimental scene: (a) Spatial sensing requirements. (b) The view from $cam4$. The red dashed circle encloses the occluded $red2$. (c) The views from $cam1$.}
\label{fig:occlusion_realworld}
\vspace{-10pt}
\end{figure}

In the \textit{Programming Phase}, we compute the sensing quality and the execution time for $red1$ and $red2$. Picking either $red1$ or $red2$ requires similar time, 28.47 and 28.34 seconds respectively. However, selecting $red1$ as the skill parameter results in good sensing quality, where $Q^{red1}_{avg}$ and $Q^{red1}_{eoi}$ are 92.27\% and 100.0\% respectively, because it is almost fully covered by $cam1$, $cam2$, and $cam3$. On the other hand, choosing to pick up $red2$ has very low sensing quality, where $Q^{red2}_{avg}$ and $Q^{red2}_{eoi}$ are 11.19\% and 0\% respectively. The zero coverage of $Q^{red2}_{eoi}$ is caused by the occlusion from the robot arm, as shown in Figure~\ref{fig:occlusion_realworld} (b).

\begin{figure}[t!]
\centering
\includegraphics[width=0.42\textwidth]{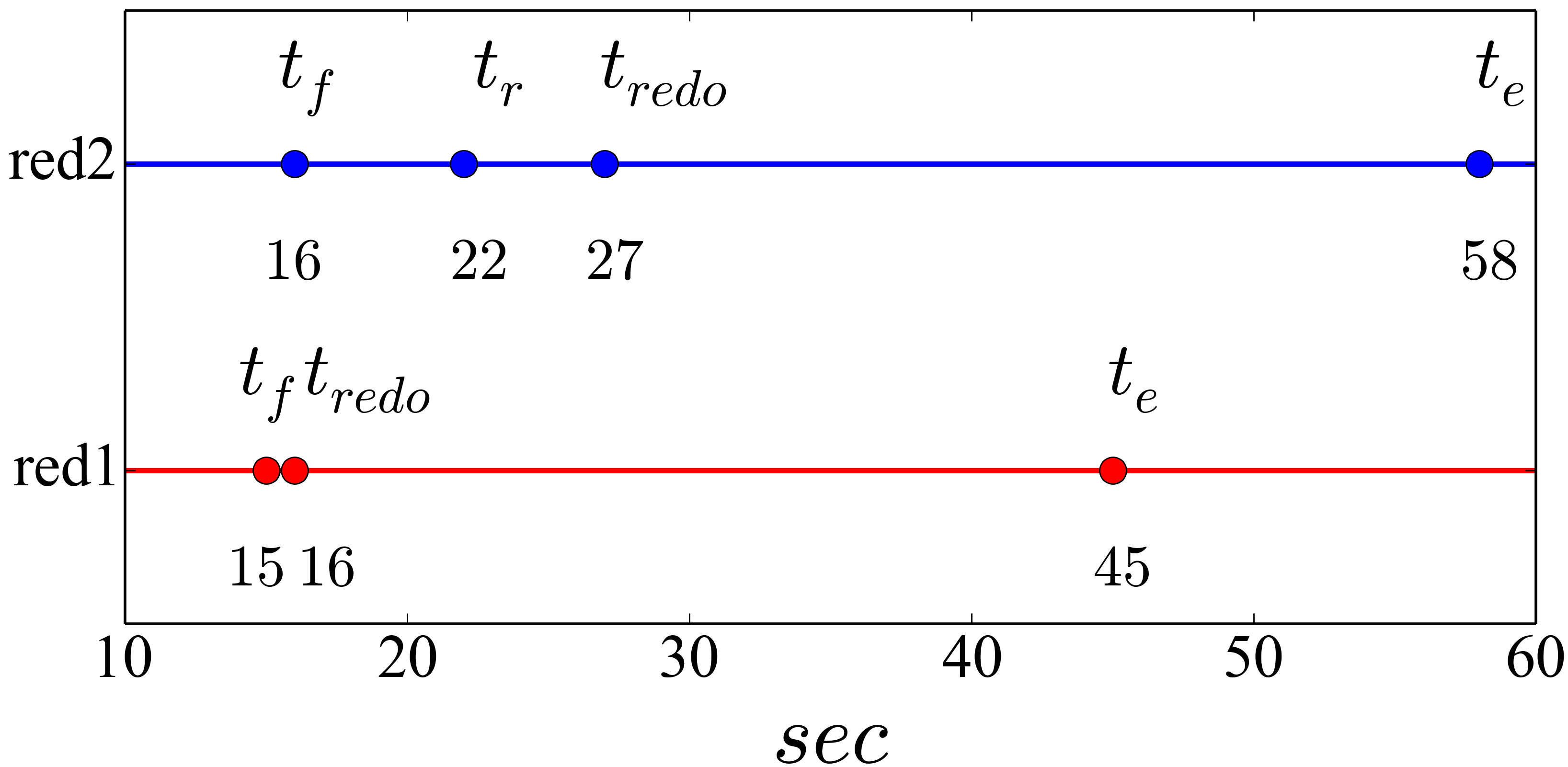}
\caption{$t_f$ is the time point when the fault, slip, occurs. $t_r$ is the time point when we detect that the object is mistakenly on the table. $t_{redo}$ is the time point when we collect enough information to decide to retry the task. $t_e$ is the time point when the skill execution is completed after the retry.}
\label{fig:demo_data}
\vspace{-15pt}
\end{figure}

We also evaluate the precomputed sensing quality with the actual detection for each skill state. Since we know the sampling time for each skill state, we examine the detection accuracy from the corresponding timestamps of the recorded video in our \textit{Runtime Phase} as our ground truth. The only time that the system fails to detect the targets while it is supposed to be covered is when the robot arm moves close to the top of $red1$, as shown in Figure~\ref{fig:occlusion_realworld} (c) from the $cam1$'s perspective. One of the ArUco markers fails to be detected probably because of the lighting condition in the experiment. For picking up $red2$, the system is able to detect the correctness of all skill states.

To show the benefit of selecting a good skill parameter, we create a scenario where the robot slips the target red screw while it is lifting it up and then our robot performs a retry. We program the robot to perform a retry only when our robot can detect that the object is mistakenly on the table and the robot gripper is functioning correctly. Figure~\ref{fig:demo_data} shows our runtime results. The results show the benefits of having better coverage that allows earlier fault detection as the robot finishes the retry $13$ seconds earlier. The data shows the advantage of incorporating coverage as the sensing quality and exposing it in the programming phase so as to reduce execution time and increase system throughput.


\section{Related Work}\label{sec:relatedwork}
Skill-based programming has been widely studied in robotic systems because it facilitates reusability~\cite{pedersen2016robot,jeong2020self}. Programmers are often given flexibility to choose different skills and parameters based on their preferences~\cite{wildgrube2019semantic,kuhner2018closed}. However, most of the extant works do not pay attention to the impact on the effectiveness of the skill monitoring modules when different skill parameters are admissible. 

Robot skills often include monitor modules to ascertain correct skill execution~\cite{pedersen2016robot}. The monitor modules usually get inputs from sensors, e.g., cameras, and perform critical event detection based on machine learning algorithms~\cite{inceoglu2018failure}. However, these works usually assume that sensors are located at the right locations that cover robot motion adequately. To know if the camera setup is actually sufficient for the current robot tasks, we incorporate camera coverage as a sensing quality and expose it in the programming phase. 

Linear Temporal Logic (LTL)~\cite{pnueli1977temporal} is a formal specification language that can be used to specify temporal properties for robot motions or tasks~\cite{he2019efficient} \cite{pacheck2020finding}. Instead of focusing on verifying  correctness of robot tasks, we focus on the sensing requirements that are extracted from temporal logic formulas that specify the robot task. 

Describing 3D locations of target objects has been widely studied in the area of active sensing in robot applications~\cite{misra2016tell}. Enclosing target objects in 3D bounding boxes is an intuitive way to describe target locations~\cite{hsieh2019lasso}. Therefore, in our work, 3D bounding boxes that enclose targets are used to define spatial sensing requirements.

\section{Conclusion}\label{sec:conclusion}
This paper presents the SQRP system which computes two sensing quality metrics, as defined by two types of average camera coverage that are used in the robot task programming phase to assist non-expert programmers to select a proper skill parameter setting for the robotic task. We use a \textit{Robot Knowledge} module to encode the robot's knowledge of the operational environment and the sensing requirements of the skill definitions. Temporal sensing requirements are expressed in Metric Interval Logic formulas to prescribe what the skill monitor system monitors and when to monitor. Spatial sensing requirements are prescribed by using 3D bounding boxes, relative poses and the distance between the target objects and the cameras. By evaluating the camera configurations in the operational environment, the SQRP system can compute the sensing qualities and provide the programmer with feedback in the programming phase. We deploy our system in both simulation and a real-world environment to obtain experimental results. We present the performance results and show that exposing the sensing quality in the programming phase can have significant benefits, both in optimizing execution time to meet run-time deadlines and in detecting run-time faults to determine if the robotic system needs to redo a sub-task.

\vspace{20pt} 

\tiny
 \bibliography{myreference}{}
\bibliographystyle{IEEEtran}

\end{document}